\pdfoutput=1
\documentclass[letterpaper]{article}
\usepackage{aaai}
\usepackage{times}
\usepackage{helvet}
\usepackage{courier}
\usepackage{helvet}  
\usepackage{courier}  
\usepackage[hyphens]{url}  
\usepackage{graphicx} 
\usepackage{array}
\usepackage{latexsym,bm,amsmath,amssymb}
\usepackage{natbib}  
\usepackage{caption} 
\DeclareCaptionStyle{ruled}{labelfont=normalfont,labelsep=colon,strut=off} 
\usepackage{algorithm}
\usepackage{algorithmic}
\usepackage{indentfirst}
\usepackage{newfloat}
\usepackage{listings}
\begin{document}
%
\title{The Brain-Inspired Decoder for Natural Visual Image Reconstruction }
\author{ Wenyi Li, Shengjie Zheng, Yufan Liao, Rongqi Hong, Weiliang Chen, Chenggnag He, Xiaojian Li\textsuperscript{\rm *}\\
    \textsuperscript{\rm 1}University of Chinese Academy of Sciences, Beijing, China\\
    \textsuperscript{\rm 2}The Brain Cognition and Brain Disease Institute (BCBDI), Shenzhen Institute of Advanced Technology, Chinese Academy of Sciences; \\
    \textsuperscript{\rm 3}Shenzhen-Hong Kong Institute of Brain Science-Shenzhen Fundamental Research Institutions, Shenzhen, China.\\
    \textsuperscript{\rm 4}CAS Key Laboratory of Brain Connectome and Manipulation\\
    \textsuperscript{\rm 5}Chengdu University of Traditional Chinese Medicine, Chengdu, China\\
    {wy.li, sj zheng, xiaojian li}@siat.ac.cn\\
}
\bibliographystyle{aaai}
\maketitle

\begin{abstract}
	\begin{quote}
 \hspace{0.2cm}Decoding images from brain activity has been a challenge. Owing to the development of deep learning, there are available tools to solve this problem. The decoded image, which aims to map neural spike trains to low-level visual features and high-level semantic information space. Recently, there are a few studies of decoding from spike trains, however, these studies pay less attention to the foundations of neuroscience and there are few studies that merged receptive field into visual image reconstruction. In this paper, we propose a deep learning neural network architecture with biological properties to reconstruct visual image from spike trains. As far as we know, we implemented a method that integrated receptive field property matrix into loss function at the first time. Our model is an end-to-end decoder from neural spike trains to images. We not only merged Gabor filter into auto-encoder which used to generate images but also proposed a loss function with receptive field properties. We evaluated our decoder on two datasets which contain macaque primary visual cortex neural spikes and salamander retina ganglion cells (RGCs) spikes. Our results show that our method can effectively combine receptive field features to reconstruct images, providing a new approach to visual reconstruction based on neural information. 
	\end{quote}
\end{abstract}

\begin{figure*}[h]
\centering
\includegraphics[width=0.4\textwidth]{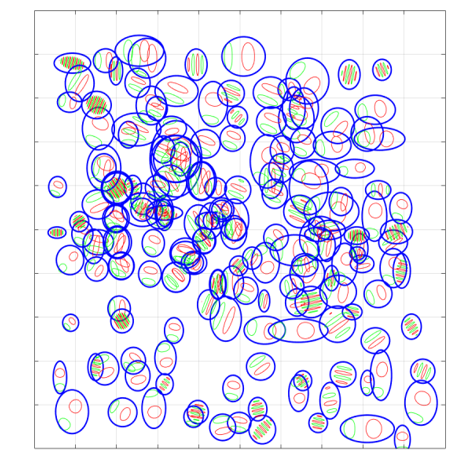} 
\caption{shows the macaque receptive fields of primary visual cortex. Each blue colored circle is 	an 	outline of receptive field. The red and green region represent excitatory and inhibitory 	subregions individually.}
\label{fig1}
\end{figure*}
\begin{figure*}[h]
\centering
\includegraphics[width=0.8\textwidth]{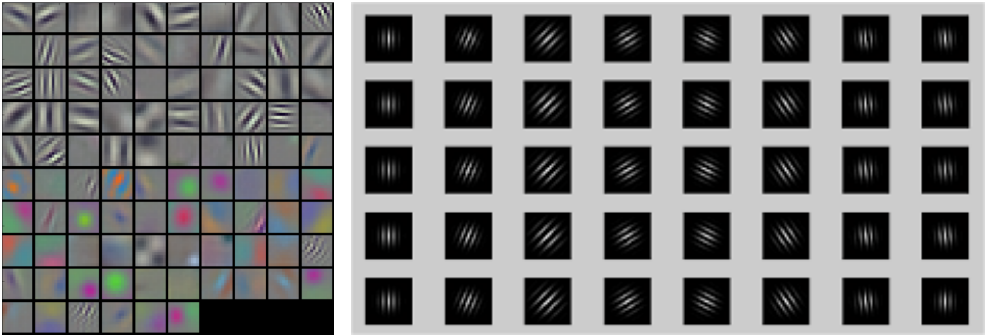} 
\caption{Research has demonstrated that filters are often redundantly learned in CNN and the most fundamental filter can be replaced by the Gabor filter. In order to compress the model with a reduced number of filter parameters, we used Gabor filters as the learned convolution filters.}
\label{fig2}
\end{figure*}
\begin{figure*}[h]
\centering
\includegraphics[width=0.5\textwidth]{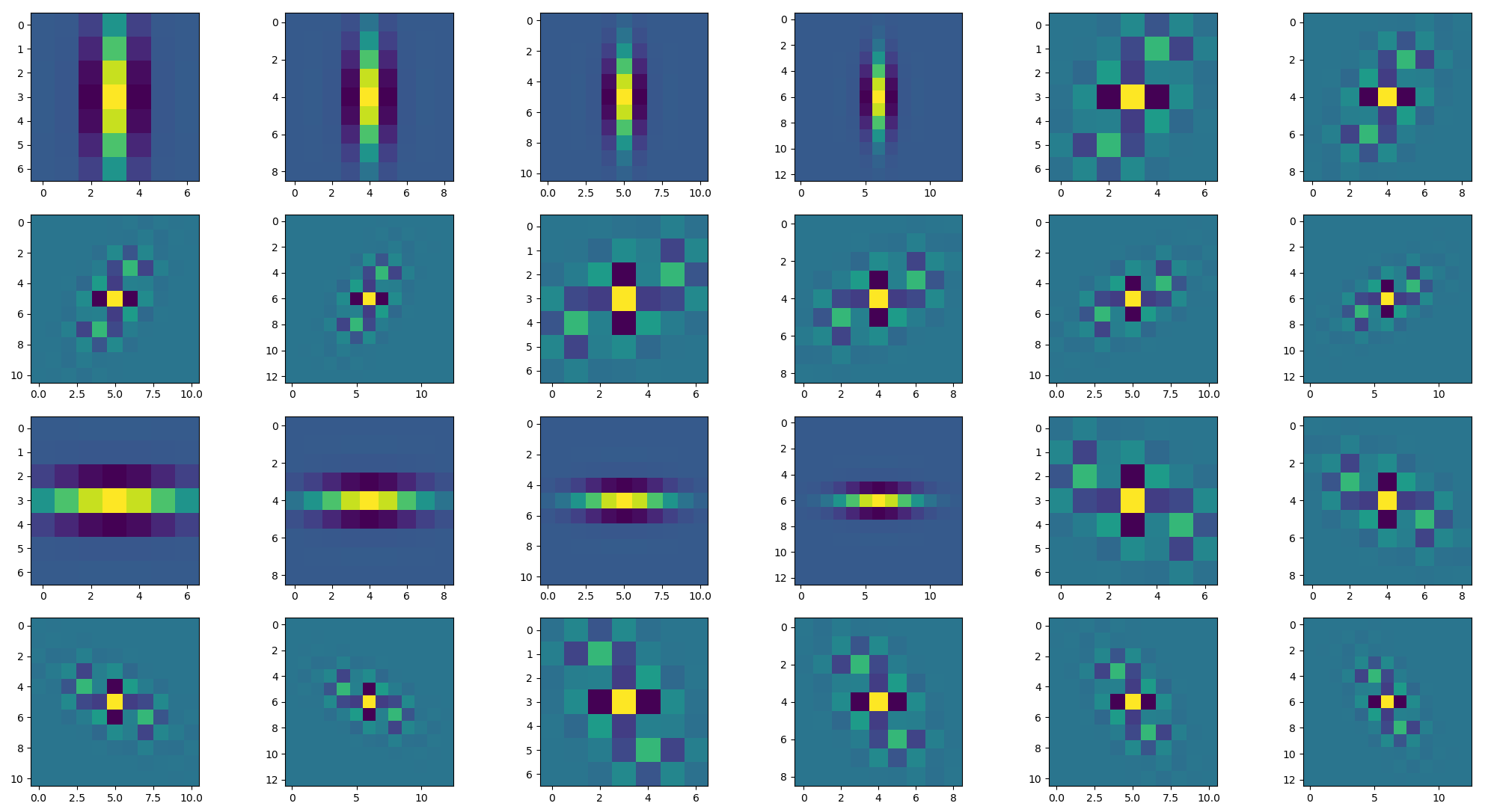} 
\caption{This illustrates shows Gabor filters with different directions and scales.}
\label{fig3}
\end{figure*}
\begin{figure*}[t]
\centering
\includegraphics[width=1\textwidth]{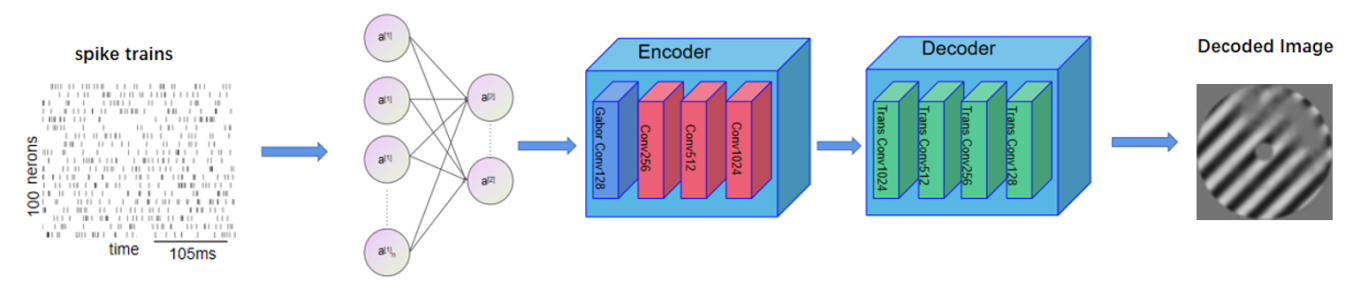}
\caption{shows the network structures of our model. We adopt Max-pooling to downsampling in the encoder and adopt transposed convolution in a decoder, which is an efficient strategy to perform upsampling. We assume ReLU as an activation function after convolution layers, the dropout is used to avoid over-fitting.}
\label{fig4}
\end{figure*}
\begin{figure*}[t]
\centering
\includegraphics[width=0.6\textwidth]{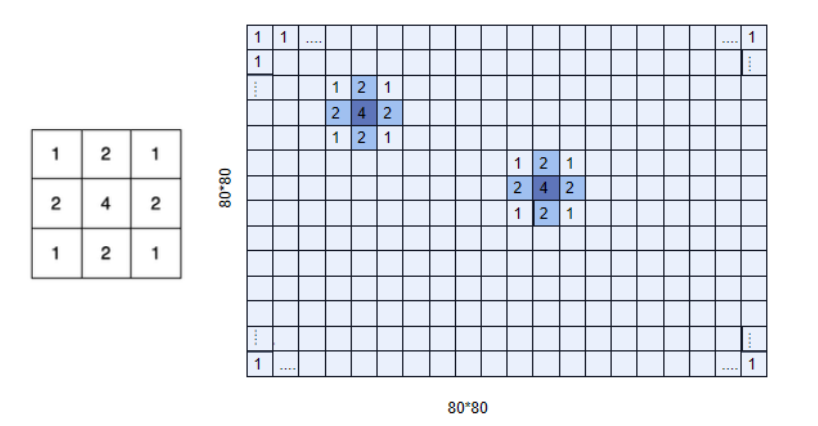} 
\caption{Left illustrates a Gaussian kernel. Right presents the W matrix which is presented the importance of the different regions of an image. }
\label{fig5}
\end{figure*}
\begin{figure*}[t]
\centering
\includegraphics[width=0.6\textwidth]{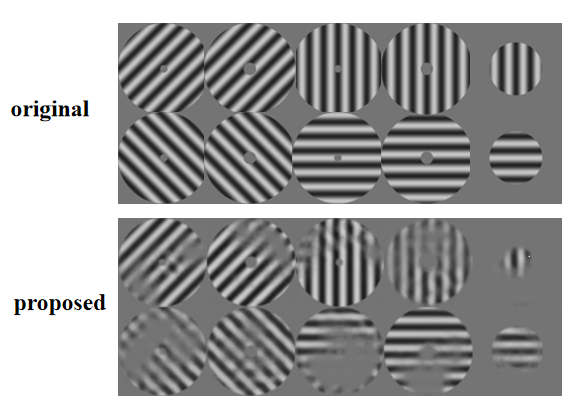} 
\caption{shows the performance of our method to reconstruct the grating images from macaque monkey V1 spike trains.}
\label{fig6}
\end{figure*}
\begin{figure*}[h]
\centering
\includegraphics[width=0.8\textwidth]{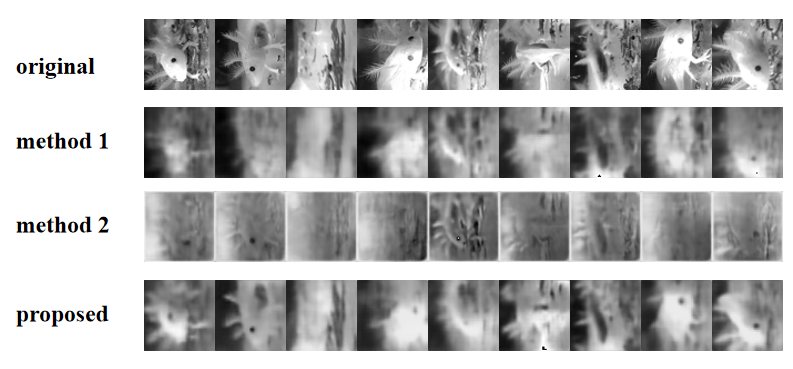} 
\caption{illustrates the Reconstructed video frames from experimental RGC spike Trains with different methods.}
\label{fig7}
\end{figure*}

\section{Introduction}
Brain-computer interface technology has made remarkable achievements in recent decades. Decoding useful information from complex brain activity is an active research topic\cite{rubin2017decoding}\cite{kay2008identifying}, which is conducive to understanding the human brain encoding and decoding mechanisms. On the other hand, this valid decoder would have great scientific and practical utility. There is a lot of research on how to read what they see based on brain activity\cite{miyawaki2008visual}\cite{nishimoto2011reconstructing}\cite{gaziv2022self}. 

In particular the visual system may consist of parallel hierarchical sequences, which are specialized for a particular task\cite{grill2004human}. Our brain can process various information from various types of receptor, and then process to generate Electrophysiological signal. Therefore, neurons and neuronal network which make up of by neurons are essential for transferring information. The interconnecting neurons process visual stimuli from the lateral geniculate nucleus to visual cortex. The visual cortex has six layers and it’s the beginning of the recognizing of the brain\cite{yucel2003effects}. Perception is processed and color and shape of an object are perceived in the several cortical areas in the parietooccipital and temporo-occipital visual association pathways\cite{chen1998human}. The primary visual cortex (V1), which is the beginning of the ventral stream and dorsal stream is located on the medial aspect of the occipital lobe. The ventral stream and dorsal stream are associated with recognition, representation and storage of long-term memory\cite{kaas2019evolution}\cite{almeida2010role}. As the brain’s primary source of information, vision is the focus of study.  How to build a general visual decoder and is it possible to reconstruct observed images from brain activity? To ensure that the visual decoder have a wide range of applications, it must be possible to decode novel images. From a task perspective, visual decoding from spike trains was mainly focused on the classification task\cite{ibos2017sequential}\cite{sachs2015brain} in the early days. but with the development of decoding algorithm, research began to shift to the field of image reconstruction; From a method perspective, many studies have focused on the application of functional magnetic resonance imaging (fMRI) in visual decoding\cite{shen2019deep}\cite{huang2021deep}\cite{horikawa2017generic}, but this is the appearance of nerve cell communication. Recently, some studies have shown that it’s possible to reconstruct image from spike trains\cite{hayashi2018image}\cite{li2022fusion}\cite{ran2021deep}. Although there are many methods to decode the brain signal, it is still difficult to decode the spike signal in the V1 brain region. Moreover, current methods lack biological basis. Therefore, we proposed to a new method with biological theory not only on neural network architecture but also loss function in deep neural network(Figure\ref{fig1}).\\

In this experiment, raster images with relatively simple structure were used for reconstruction, and spike trains of macaque V1 were used as input data. We used an end-to-end neural network to decode the spike trains and the output is the reconstructed images. The realization of this task provides a reference for image reconstruction with more complex structures.\\

In order to test our decoder on different spike datasets, We tested the decoder on RGCs spike trains, and we found that it can obtain good performance in the datasets of RGCs for the reconstruction of natural scenes of video frames.
\section{Material and Method }
\subsubsection{Implementation}
We implemented the proposed algorithm on 3070ti GPU, the i5 9400F CPU, and 64G RAM. To raise computing speed, we read data from the hard disk to memory at first.. Our code is based on the environment torch1.9.1 and python 3.8. The version of Cuda is 11.4. The code and results are available at https://github.com/WYCAS/S2INet
\subsection{Datasets}
When information on the spike trains is insufficient to reconstruct a high-resolution image, the weighted loss function can reconstruct images discriminately, where the receptive field region has a higher weight.
\subsubsection{Macaque primary cortex datasets}
We tested our model on two natural neural spike trains. The first datasets are macaque V1 datasets which consist of multi-electrode recordings from V1 in anesthetized macaque monkeys, while natural images and gratings were flashed on the screen. The data were collected in the Laboratory of Adam Kohn at the Albert Einstein College of Medicine and downloaded from the CRCNS website. Data was recorded by the "Utah" electrode array. Natural images were presented at two sizes, 3-6.7 degrees and windowed to 1 degree, to quantify surround modulation. The receptive field(RF) was measured using small gratings presented at a range of positions. The RF center of each neuron was defined as the location of the peak of a 2D Gaussian fit to the spatial activity map. Experimental procedures and stimuli are fully described in the associated paper\cite{coen2015flexible}.
\subsubsection{Salamander retina datasets}
The second datasets are temporal firing patterns in populations of simultaneously recorded salamander neurons. Our decoder was used to reconstruct video frames from the spike trains of a population of RGCs of salamanders. Visual stimuli were projected onto the retina through a telecentric lens. These datasets and descriptions can be found in this paper\cite{onken2016using}. The video frame data consists of salamander retina spike trains which include 1800 video frames, and resize the pixels to 64*64 pixels. There are 1800 video frames as stimulus and the number of 49 RGCs spike trains. The training set contains 1440(1800*0.8) video frames of 64*64 pixels, the test set contains 360(1800*0.2) video frames.
\subsection{Data Process}
In the macaque V1 datasets, we used 7 session data that included the results of multiple experiments. We used the 80$\%$ data(1249) as the training set and the remaining 20$\%$ data as the test set. To ensure that each batch input dimension is equal, we intercept a small part of the data. The data of 100 neurons in 105ms were retained every session. The reconstructed image is modified to 80*80 pixels grayscale. And as to the population of RGCs spike trains in the retinas of salamanders, we converted the timestamp to spike trains on a 10ms scale. We randomly selected some scenes and disrupt the order to wash out the temporal correlation within the video.
\subsection{Model}

In this experiment, the structure of the auto-encoder is used\cite{hinton2006reducing}. In addition, considering the directivity of raster images and the training complexity of CNNs, the first convolution layer of our model used the Gabor filters\cite{luan2018gabor}(Figure\ref{fig2}).

Gabor convolution neural network is a deep neural network using Gabor orientation filters (GoFs), which can produce feature map to enhance directions and scales information(Figure\ref{fig3}). In addition, GoFs are wildly used to model receptive fields of simple cells of the visual cortex. In this way, the model of deep learning can be strengthened while learning fewer parameters. Applying the convolutional neural network to the raster image reconstruction task can better fit the raster image with directions and scales. We adopted the following neural network framework(Figure\ref{fig4}).
\subsection{Loss function}

We designed a new loss function and set receptive field properties of the weight matrix, which can be adjusted according to the position of the receptive field. Applying the weight matrix of the receptive field to the loss function can make the different definitions in different parts of the reconstructed image and give higher weight to the area of attention of the receptive field. Due to the need to measure the structural similarity between the original image and the reconstructed image, we fuse the Structure Similarity Index Measure (SSIM) in the loss function. SSIM can be described in detail in this paper\cite{wang2004image}. It can be summarized as follows: The comparison measurements are luminance, contrast, and structure. they are described as:\\
\begin{equation}
l_{luminace}(x,y) = \frac{2\mu_{x}\mu_{y} + c_{1}}{\mu_{x}^{2} + \mu_{y}^{2} + c_{1}}
\end{equation}
\begin{equation}
c_{contrast}(x,y) = \frac{{2\sigma}_{x~}\sigma_{y} + c_{2}}{\sigma_{x}^{2} + \sigma_{y}^{2} + c_{2}}
\end{equation}
\begin{equation}
s_{structure}(x,y) = \frac{\sigma_{xy} + c_{3}}{\sigma_{x}\sigma_{y} + c_{3}}
\end{equation}
\begin{equation}
SSIM = \left\lbrack l(x,y)\rbrack^{\alpha}\left\lbrack c(x,y)\rbrack^{\beta}\left\lbrack s(x,y)\rbrack^{\gamma} \right. \right. \right.
\end{equation}
$\mathrm{\mu_{x}}$ is the mean of $\mathrm{x}$, $\mathrm{\sigma_{x}}$ is the standard deviation of $\mathrm{x}$, and $\mathrm{\sigma_{xy}}$ is the covariance between $\mathrm{x}$ and $\mathrm{y}$. Generally. We set $\mathrm{\alpha}$, $\mathrm{\beta}$, and $\mathrm{\gamma}$ to be 1, and ${c_3}$ =${c_2}$/2.\\ The SSIM can be described as:
\begin{equation}
SSIM(x,y) = \frac{\left( 2\mu_{x}\mu_{y} + c_{1} \right)\left( {2\sigma}_{xy} + c_{2} \right)}{\left. \left( \mu \right._{x}^{2} + \mu_{y}^{2} + c_{1} \right)\left( \sigma_{x}^{2} + \sigma_{y}^{2} + c_{2} \right)}
\label{Equa5}
\end{equation}
\begin{equation}
MSE = \frac{1}{I_{h} \times I_{w}}{\sum\limits_{i = 1}^{H}{\sum\limits_{j = 1}^{W}\left( X_{1}(i,j) - X_{2}(i,j))^{2} \right.}}
\end{equation}
\begin{equation}
L = {\mu L}_{SSIM} + (1 - \mu){W}{L}_{MSE}
\end{equation}
$\mathrm{W}$ is a matrix with receptive field properties \ref{Equa5}. We used a Gaussian kernel as the spatial weight matrix.\\
\section{Result}
The performance of our method was evaluated on two open source datasets which include macaque primary motor cortex and salamander retina spike trains. We didn’t use a simulator, we chose real datasets for training our model. We evaluated our method on images of gratings that contain 4 orientations and a different degree in diameter. Figure\ref{fig6} shows the reconstruction effects of our method. Due to the location of the receptive field, there are different reconstruction effects in the different regions. \\
To further test the generalization capability of our method, we did experiments on biological experimental RGCs data to RGCs responses(Figure\ref{fig7}). The first method (method 1) is based on an auto-encoder with an MSE loss function and the second method(method 2) is based SSIM loss function\cite{zhang2020reconstruction}.\\
\begin{table}[htbp] 
		\setlength{\abovecaptionskip}{0cm} 
		\setlength{\belowcaptionskip}{0.2cm}
	\centering
	\caption{\label{tab:l1}shows the performance of our method compared with other methods} 
	\begin{tabular}
	{p{1.5cm}<{\centering}p{1.1cm}<{\centering}p{1.1cm}<{\centering}p{1.1cm}<{\centering}p{1.1cm}<{\centering}}
		\hline
		Method  &MSE    &PSNR   &VIFP   &SSIM\\ 
		 \hline
		Method 1&1.8611& 9.3023&0.0554&0.7281\\ 
		Method 2&0.5455&14.6389&0.2183&0.8088\\  
		\textbf{Proposed}&\textbf{0.5147}&\textbf{14.8911}&\textbf{0.2316}&\textbf{0.8174}\\ 
		\hline
	\end{tabular}
\end{table} 
Table\ref{tab:l1} shows the performance of our method compared with the CNN auto-encoder with the mean square error (MSE), peak signal to noise ratio (PSNR), visual information fidelity, pixel domain version (VIFP)\cite{han2013new}, and SSIM loss function. MSE describes the absolute difference of every pixel, the PSNR describes the global quality, the PSNRis defined as:
\begin{equation}
PSNR = 10 \cdot {log}_{10}\left( \frac{L^{2}}{MSE} \right)
\end{equation}

L presents the maximum pixel value(255 for 8-bit images).
The VIFP that quantify the information shared between the test and the reference images, and the SSIM that captures the structure similarity, for evaluating the reconstruction results. Comparing other method our model suggests that Gabor auto-encoder architecture with weighted loss function enables more clear reconstruction. The proposed method can do well in presenting the reconstruction details, especially in complex stripe feature of animal or scenes. However,its performance is still poor for stimuli from complex visual images, This may be due to short stimuli and complex feature of each natural image.
\section{Discussion}

In this study, we propose a new reconstruction framework that can reconstruct images based on neural spike trains in the visual cortex and retina. We build an end-to-end model from spike trains to images that can be used for various data sets. The reconstruction module has a simple encoding and decoding structure, which extracts information from the spike train. We propose a Gabor autoencoder and a new loss function that integrates the properties of the receptive field, the Gabor filter is used to represent the retinal processing mechanism of the spike train. We argue that the effect of image reconstruction from macaque V1 spike trains and salamander retinal spike trains is different between them due to the complexity of neural encoding. Depending on the type of neural information, the image reconstruction effect is related to the proportion of effective neuronal information collected by the neural electrodes.

In future work, we will further improve the model to accommodate complex texture images. The decoder can be used to explore the relationship between the number of neurons and the reconstructed image. And, we can also consider the problem of visual reconstruction based on continuous visual neural signals, which is due to the logical continuity of visual information, which combines the coding characteristics of continuous information as well as discrete information, and we may be able to reconstruct the image better if we consider the historical information of vision as well as the current stimulus.

Although the decoding model can reconstruct a part of the visual information, the model cannot yet theoretically establish the decoding mechanism of vision. Based on the data-driven approach of deep learning, we also expect that in the future, using more reliable data, we will be able to effectively decode visual information, and then invert the encoding mechanism of vision, and this visual encoding mechanism can be directly used in brain-inspired algorithms to achieve brain-inspired intelligence with biological interpretability.

\bigskip
\bibliography{aaai}
\end{document}